\documentclass{article}

\usepackage{PRIMEarxiv}

\usepackage[utf8]{inputenc} 
\usepackage[T1]{fontenc}    
\usepackage{hyperref}       
\usepackage{url}            
\usepackage{booktabs}       
\usepackage{amsfonts}       
\usepackage{nicefrac}       
\usepackage{microtype}      
\usepackage{lipsum}
\usepackage{fancyhdr}       
\usepackage{graphicx}       
\graphicspath{{media/}}     

\usepackage{authblk}
\usepackage{booktabs}
\usepackage{xspace}

\usepackage{pifont}
\newcommand{\cmark}{\ding{51}}%
\newcommand{\xmark}{\ding{55}}%
\usepackage{color}
\usepackage{array}
\usepackage{amsmath}
\usepackage{threeparttable}
\usepackage{multirow}

\usepackage[linesnumbered,ruled,vlined]{algorithm2e}
\SetKwInput{KwInput}{Input}                
\SetKwInput{KwOutput}{Output}              
\SetKwInput{KwInitial}{Initialization}
\SetKwFor{For}{for}{}{}

\title{Cross-Modal Contrastive Learning for Robust Reasoning in VQA
}

\author[1]{Qi Zheng}
\author[2]{Chaoyue Wang}
\author[2]{Daqing Liu}
\author[3]{Dadong Wang}
\author[1,2]{Dacheng Tao}
\affil[1]{University of Sydney, NSW 2008, Australia}
\affil[2]{JD Explore Academy}
\affil[3]{DATA61, CSIRO, NSW 2122, Australia}

\begin{document}
\maketitle


\begin{abstract}
Multi-modal reasoning in visual question answering (VQA) has witnessed rapid progress recently. However, most reasoning models heavily rely on shortcuts learned from training data, which prevents their usage in challenging real-world scenarios. In this paper, we propose a simple but effective cross-modal contrastive learning strategy to get rid of the shortcut reasoning caused by imbalanced annotations and improve the overall performance. Different from existing contrastive learning with complex negative categories on coarse (Image, Question, Answer) triplet level, we leverage the correspondences between the language and image modalities to perform finer-grained cross-modal contrastive learning. We treat each Question-Answer (QA) pair as a whole, and differentiate between images that conform with it and those against it. To alleviate the issue of sampling bias, we further build connected graphs among images. For each positive pair, we regard the images from different graphs as negative samples and deduct the version of multi-positive contrastive learning. To our best knowledge, it is the first paper that reveals a general contrastive learning strategy without delicate hand-craft rules can contribute to robust VQA reasoning. Experiments on several mainstream VQA datasets demonstrate our superiority compared to the state of the arts. Code is available at \url{https://github.com/qizhust/cmcl_vqa_pl}.
\end{abstract}

\section{Introduction}
Although great progress has been made on visual question answering (VQA), it is observed that reasoning agents tend to learn shortcuts~\cite{agrawal2018don,kervadec2021roses}. Shortcuts are spurious correlations in training data that lead to correct answers without deploying the desirable reasoning process, e.g., guessing answers by the dominant object in an image or even without looking at the image~\cite{agrawal2018don,dancette2021beyond,kervadec2021roses}. Shortcut representations blind the agents with specific training statistics and prevent generalizing to real-world scenarios.

Recent works show that shortcuts learned by neural networks are caused by imbalanced training annotations~\cite{kervadec2021roses,goyal2017making}. For instance, ``tennis'' is the correct answer for 41\% of the ``What sport...?'' questions in VQAv1~\cite{antol2015vqa}. Manually increasing and balancing the annotations could be a choice, but it is expensive and laborious. Alternatively, the imbalance issue can be alleviated by regrouping elements among the (`Question', `Answer', `Image') triplet. For example, ``tennis'' can be a correct answer to the ``What sport...?'' question for a ``tennis'' image, but it is a \textit{wrong} answer for a ``cooking'' image. Inspired by this, ConClaT~\cite{kant2021contrast} is the first work that applies contrastive learning in VQA problems, which mainly targets the issue of robustness to linguistic variations. It augments training data with generated question paraphrases and designs three types of negative samples to compose contrastive pairs, e.g., $(IQA,~I^-QA)$. Although contrastive learning is an intuitive solution for the shortcut problem and has been validated on multiple vision and language tasks, in the VQA task, the effective method has to cooperate with delicate hand-craft designs, such as cumbersome data augmentation and alternately training at specific ratio~\cite{kant2021contrast}.

In this work, we develop a finer-grained cross-modal approach that is simpler and more effective with robust visual reasoning ability. A novel cross-modal contrastive learning strategy is proposed for VQA. Concretely, each Question-Answer (QA) matching in the triplet is regarded as a whole, then a $QA$ matching with its paired image $I^+$ forms a positive $(QA,~I^+)$ pair, while a $QA$ matching with its unpaired image $I^-$ forms a negative pair. Thus, the cross-modal contrastive enables a soft alignment between vision and language semantic spaces and promotes learning the abstract common knowledge from the alignment to allow robust reasoning. Moreover, to overcome the sampling bias~\cite{chuang2020debiased} caused by the random scheme, we build connected graphs in the feature space of images to avoid false negative triplets. For example, assuming that QA is ``What color is the sky?-Blue'' for the matched image $I$, yet there is the blue sky in the background of an unpaired image $I^-$, then $(QA,~I^-)$ is a false negative pair, which undermines the learning process. By constructing connected graphs, we can take the images outside the connected graphs of $I$ as negative samples for QA matching. Jointly training with answer classification, our method enables a powerful reasoning agent to learn more abstract knowledge by distinguishing cross-modal pairs. More importantly, our work first reveals that a general contrastive learning strategy without dedicated hand-craft rules can contribute to robust VQA reasoning.

We show that in mainstream VQA datasets, our method outperforms its baseline counterparts and existing methods on difficult questions. For other categories that usually benefit from shortcuts, ours achieves comparable results. Specifically, to evaluate the efficacy of the proposed method, we first conduct experiments on the regularly balanced VQAv2 dataset. Our method achieves similar accuracy as the baseline model on \textit{Yes/No} questions where shortcuts are more likely to benefit answering and surpasses it on \textit{Number} questions where shortcuts scarcely help. Consistent results can be observed in the VQA-CE dataset. Our method shows superiority on \textit{Hard} split where no shortcuts exist and scores on a par with the baseline on \textit{Counter} and \textit{Easy} splits where either count-shortcut or shortcut leads to correct answers. Finally, we test our method on the VQA-CP dataset and find it improves reasoning performance as well. It verifies that our method works well when shortcuts in questions are explicitly reduced and robust reasoning is required. It is worth noting that the improvements above are insensitive to the strength of a baseline model. To summarize, our contributions are threefold
\begin{itemize}
    \item We propose a simple but effective way to integrate fine-grained cross-modal contrast with supervised learning for VQA. The reasoning ability is enhanced by considering the matched QA pair as a whole and differentiating among images that conform with it and those against it. 
    \item To alleviate the sampling bias issue, we build connected graphs among images and compose negative samples by those from different graphs. It avoids undermining the reasoning ability in situations when shortcuts help question answering.
    \item Our work is the first to show that a general contrastive learning strategy can benefit robust reasoning in VQA without complex hand-craft training schemes, compared with the supervised counterparts. 
\end{itemize}

\section{Related Work}
\paragraph{Visual Question Answering (VQA)}
Since open-ended VQA was proposed in~\cite{antol2015vqa} and a revised version in~\cite{goyal2017making}, the solutions have been developed for roughly two periods. The first phase is based on object detectors~\cite{ren2015faster,he2017mask} and LSTMs~\cite{hochreiter1997long,graves2013speech,zia2019long}, including those equipped with various attentions. SAN~\cite{yang2016stacked}, UpDown~\cite{anderson2018bottom}, BLOCK~\cite{ben2019block} are representatives among them. While the second phase emerges with the impressive success of transformers~\cite{vaswani2017attention}, diverse transformer-based backbones provide a general pretraining-finetuning paradigm for multi-modal tasks. The well-known backbones include ViLBERT~\cite{lu2019vilbert}, VisualBERT~\cite{li2019visualbert}, LXMERT~\cite{tan2019lxmert}, UNITER~\cite{chen2020uniter}, ViLT~\cite{kim2021vilt}, METER~\cite{dou2021empirical} etc. 

With the great improvements of overall performance in regular open-ended VQA, there are increasing works focused on VQA in more challenging or specific situations for a better deployment in real-world scenarios. \cite{kervadec2021roses} designs the GQA-OOD benchmark that compares accuracy over both rare and frequent question-answer pairs. A new setting is proposed in VQA-CP~\cite{agrawal2018don} where train and test splits have different prior distributions of answers. Besides, an evaluation methodology is introduced in VQA-CE~\cite{dancette2021beyond} to diagnose cases of shortcut learning. Different from works that target biased distribution only~\cite{cadene2019rubi,chen2020counterfactual,kervadec2021roses,niu2021counterfactual,gupta2022swapmix}, our method aims to promote overall robustness of VQA reasoning through multi-modal alignment.

\paragraph{Contrastive learning}
Recently, contrastive learning has achieved a great success in computer vision~\cite{he2020momentum,chen2020improved,chen2021empirical,wang2020understanding,chen2020simple,grill2020bootstrap,chen2021exploring} and natural language processing (NLP)~\cite{wu2020clear,yan2021consert,gao2021simcse}. It has also been adapted to vision-language multi-modal representation learning. For instance, CLIP~\cite{radford2021learning} jointly trains an image encoder and a text encoder to predict the matching of (image, text) pairs and acquires outstanding zero-shot ability. CrossCLR~\cite{zolfaghari2021crossclr} considers intra-modality similarities to learn cross-modal embeddings for video and text to improve embedding alignment. 

Existing VQA methods that involve contrastive learning~\cite{liang2020learning,kim2021self} contrast negative samples with fictitious data, e.g., false image feature synthesis~\cite{liang2020learning} and text operation or rephrasing~\cite{kim2021self}. It breaks the real data distribution and is fragile to false negatives due to imperfect object detection or question paraphrasing. ConClaT~\cite{kant2021contrast} builds coarse-level contrast and generate question paraphrases for data augmentation using Back-translation. We propose a simplex methodology that contrasts negative samples by recombining real data, and we are also the first work that focuses on the false negative issue in VQA.

\begin{figure*}[t]
    \centering
    \includegraphics[width=\textwidth]{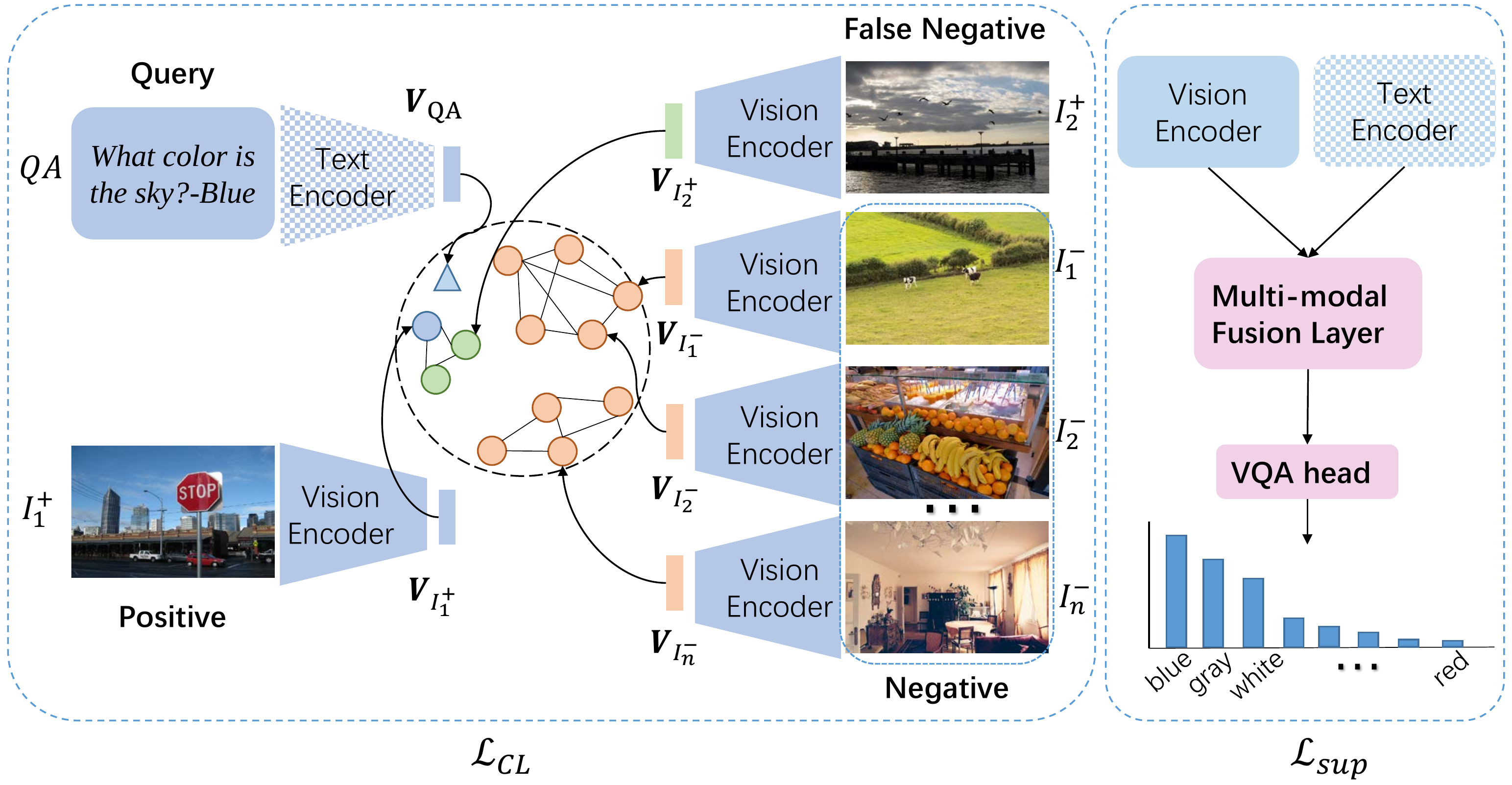}
    \caption{\textbf{Overview of the proposed cross-modal contrastive learning.} For the objective $\mathcal{L}_{CL}$, each encoded question-answer matching is treated as a whole, namely the Query, and its paired image is a positive sample. Connected graphs are built in the representation space of images within a mini-batch to reduce false negatives. The final negative images ${I_1^-,~I_2^-,\dots,I_n^-}$ are selected from those outside the graph of the positive one $I_1^+$. On the right is a typical multi-modal backbone architecture. A reasoning model is jointly trained with cross-entropy objective $\mathcal{L}_{sup}$ and contrastive objective $\mathcal{L}_{CL}$.}
    \label{fig:framework}
\end{figure*}


\section{Method}
An overview of the proposed cross-modal contrastive learning and a flowchart of supervised training is provided in Figure~\ref{fig:framework}. For the backbone architecture, a VQA reasoning agent consists of a perception module (i.e., a vision encoder \& a text encoder), a multi-modal fusion module, and an MLP head. Imbalanced annotations enable the agent to derive correct answers via shortcuts in representations learned by the perception module. Therefore, we apply contrast signals in the cross-modal hidden space to improve perception quality and allow robust reasoning. In this section, we first briefly introduce supervised training for VQA and then describe the proposed cross-modal contrastive learning method.

\subsection{Supervised Learning for VQA}

We denote the vision encoder as $f_\theta:~I\to V_I \in \mathbb{R}^{D_1}$, and the text encoder as $f_\varphi:~X\to V_{X} \in \mathbb{R}^{D_2}$, where the input $X=\{Q\}$ for supervised training; $D_1$, $D_2$ are the dimension of the multi-modal hidden spaces. We assume that the space of two modalities has the same dimension for simplicity, i.e. $D_1=D_2=D$. We treat the merged attention module and the MLP head as a whole by $g_\phi:~(V_I,~V_X)\to \mathbb{R}^K$, where $K$ is the categories of ground-truth answers. $\theta,~\varphi,~\phi$ are parameters to be learned.

Given a mini-batch of $M$ (image, question, answer) training triplets, i.e., $\mathcal{D}=\{(I,~Q,~A)\}_1^M$, supervised training with cross-entropy loss minimizes the following objective,
\begin{align} \label{eq:sup}
    \mathcal{L}_{sup}=-\frac{1}{M}\sum_{m=1}^M \log P\left(g_\phi(f_\theta(I_m),f_\varphi(Q_m)){=}A_m\right).
\end{align}

\subsection{Cross-Modal Contrastive Learning for Robust Reasoning}
\paragraph{Contrastive learning for VQA.}
As analyzed previously, contrastive learning can be exploited to alleviate shortcuts learned from imbalanced training data. Intuitively, setting a matched $(I,~Q,~A)$ as a query, positive samples can come from augmented triplets $(I,~Q',~A)$ or $(I',~Q,~A)$, and negative samples can be obtained by replacing $I$ or $Q$ with mismatched ones. However, we argue that two shortcomings exist in the vanilla solution. On the one hand, the contrast of positive pairs, e.g., $(IQA,~IQ^+A)$ and negative pairs, e.g., $(IQA,~IQ^-A))$, are at a relatively coarse level. In preliminary experiments, numerous augmented samples or subtle techniques would be required to take effect. On the other hand, image data augmentation has been conventionally used in supervised training and would incur a cumbersome data augmentation process as in ConClaT~\cite{kant2021contrast}.

\paragraph{Cross-modal contrastive learning.} 
To extricate the dilemma, we introduce contrasts between the language and visual modalities. Instead of performing contrastive learning on the coarse triplet level (e.g., $(IQA,~IQ^+A)$), in our method, each Question-Answer (QA) matching in VQA is regarded as a whole, then a $QA$ matching with its paired image $I^+$ forms a positive $(QA,~I^+)$ pair, while a $QA$ matching with its unpaired image $I^-$ forms a negative pair. Originally, negative samples $\{I_1^-,I_2^-,\dots,I_n^-\}$ come from all the images within the mini-batch except the matched $I$, and the basic InfoNCE loss~\cite{oord2018representation} minimizes
\begin{align}\label{eq:base}
    \mathcal{L}_{CL}=-\frac{1}{M}\sum_{m=1}^M \mathcal{J}_{CL}^{(m)} =-\frac{1}{M}\sum_{m=1}^M \log \frac{e^{\left(h_\eta (V_{{(QA)}_m},V_{I_m})/\tau\right)}}{\sum_{j=1}^M e^{\left(h_\eta (V_{{(QA)}_m},V_{I_j})/\tau\right)}} \doteq -\frac{1}{M}\sum_{m=1}^M \log \frac{e^{\left(h_\eta^{(mm)}/\tau\right)}}{\sum_{j=1}^M e^{\left(h_\eta^{(mj)}/\tau\right)}},
\end{align}
where $V_{(QA)_m}{=}f_\varphi(Q_m,~A_m)$ during training, $h_\eta$ estimates the similarity of two samples, and $\tau$ is the temperature parameter. In this paper, we use cosine similarity for the estimation, and a linear transformation is added to align the dimension of inputs.

\paragraph{Building connected graphs for negative sample selection.} In practice, we observed false negative samples created by the vanilla negative sampling way. As demonstrated in Figure~\ref{fig:framework}, false negative samples interfere with reasoning in regular situations. The image $I_2$ is treated as a negative sample in the vanilla selection strategy, which is incorrect. Therefore, we build connected graphs in $V_I$ space and take the images outside the connected components of $I_m$ as unpaired images for the QA matching $(QA)_m$.

We construct a graph $\mathcal{G}=(\mathcal{V},~\mathcal{E})$ where nodes $\mathcal{V}\in \{V_{I_1},V_{I_2},\dots,V_{I_M}\}$ and the edge $\mathcal{E}=\{e_{ij}\},~0\leq i,j\leq M$. First, we find the nearest neighbor for each node,
\begin{align}\label{eq:edge}
    e_{ij}=\left\{ \begin{array}{ll}
        1 & \textrm{if $j=\arg\min_k d(V_{I_i},V_{I_k})$} \\
        0 & \textrm{otherwise}
    \end{array} \right.
\end{align}
where we use cosine similarity as the negative distance for the measurement, i.e., $d(V_{I_i}, V_{I_k})=-\frac{V_{I_i}^T V_{I_k}}{\parallel V_{I_i}\parallel \parallel V_{I_k}\parallel}$. Eq.~(\ref{eq:edge}) gives us a sparse and symmetric 1-nearest neighbor graph. To identify potential false negative samples, we find more similar samples for each node by building connected components among them. This problem can be efficiently solved by the Hoshen–Kopelman algorithm~\cite{hoshen1976percolation}.

We denote the derived connected components (i.e. fully-connected sub-graphs ) as $\mathcal{G}_i=(\mathcal{V},\mathcal{E})$, where $\mathcal{V}\in \{V_I\}^i$, $\mathcal{E}=[1]_{|V_I^i|\times |V_I^i|}$ and $1\leq i\leq M$. Then for a node from $\mathcal{G}_i$, its final negative samples $V_{I^-}$ comes from all the connected components $\mathcal{G}_k~(k\neq i)$. We deduct the first refined cross-modal contrastive learning objective as
\begin{align}\label{eq:neg}
    \mathcal{J}_{CL}^{(m)} = \log \frac{e^{\left(h_\eta^{(mm)}/\tau\right)}}{e^{\left(h_\eta^{(mm)}/\tau\right)} {+} \sum_{k=1,k\neq i}^{|\mathcal{G}|}\sum_{V_{I_j}\in \mathcal{G}_k} e^{\left(h_\eta^{(mj)}/\tau\right)}}, 
\end{align}
where $V_{I_m}\in \mathcal{G}_i$ and $|\mathcal{G}|$ is the number of total sub-graphs.

It is worth noting that Eq.~(\ref{eq:neg}) only involves negative samples and ignores the similar samples in $\mathcal{G}_i$. By using those positive samples, we can derive the fully multi-positive (MP) version objective as
\begin{align}\label{eq:mp}
    \mathcal{J}_{CL}^{(m)} = \frac{1}{|\mathcal{G}_i|}\sum_{V_{I_c}\in \mathcal{G}_i} \log \frac{e^{\left(h_\eta^{(mc)}/\tau\right)}}{\sum_{k=1}^{|\mathcal{G}|}\sum_{V_{I_j}\in \mathcal{G}_k} e^{\left(h_\eta^{(mj)}/\tau\right)}},
\end{align}
where $|\mathcal{G}_i|$ is the number of vertices in $\mathcal{G}_i$, which is determined by the matching image $I_m$ for $(QA)_m$.

Combining supervised training with contrastive learning, we minimize the sum of two losses
\begin{align}\label{eq:joint}
    \mathcal{L} = \mathcal{L}_{sup} + \lambda \mathcal{L}_{CL} = \mathcal{L}_{sup} - \frac{\lambda}{M}\sum_{m=1}^M \mathcal{J}_{CL}^{(m)},
\end{align}
where $\lambda$ is a weighting hyperparameter. $\mathcal{L}_{sup}$ is calculated by Eq.~(\ref{eq:sup}) and $\mathcal{J}_{CL}^{(m)}$ by Eq.~(\ref{eq:mp}).

\subsection{Training and Discussions}
ConClaT~\cite{kant2021contrast} concludes that contrastive learning is merely effective by alternate training, which incurs at least one extra sensitive hyperparameter, namely the alternate ratio. However, by designing a finer-grained contrastive learning method, we empirically show that the simple joint training in Eq.~(\ref{eq:joint}) can improve reasoning ability and alleviate the shortcut issue.

The \textit{semantic collision} issue identified in CrossCLR~\cite{zolfaghari2021crossclr} is also known as \textit{class collision} or \textit{sampling bias}. Interestingly, the solution, i.e., influential sampling and negative set pruning, proposed in CrossCLR~\cite{zolfaghari2021crossclr} can be seen as a particular case of our method, where a rigid threshold is set to choose positive samples. By constructing connected components, we can not only transfer between a single positive sample and fully multi-positive samples but also determine how many connected components involve in the calculation, which is more flexible than setting a threshold.

For the proposed method, we treat the merged attention module and the MLP head as a whole and optimize the feature space of the vision encoder and the text encoder by cross-modal contrastive learning. Alternatively, we can view the two encoders and the merged attention module as a whole and optimize the output space of the merged attention module. We leave out this for open explorations.

\section{Experimental Results}

\subsection{Datasets and Implementation Details}\label{subsec:imple}
VQAv2\footnote{https://visualqa.org/download.html}~\cite{goyal2017making} is constructed upon the real-image VQA dataset in~\cite{antol2015vqa} to counter its strong language biases. It contains about 443K, 214K, and 453K instances for training, validation, and testing, respectively. For every (image $I$, question $Q$, answer $A$) triplet in VQAv2, it is likely that ($I'$, $Q$, $A'$) exists where $I'$ is similar to $I$ but $A'$ is different from $A$. To some degree, this provides a relatively balanced training and inference setting. To examine reasoning ability in more challenging scenarios, we conduct more experiments on VQA-CE\footnote{https://github.com/cdancette/detect-shortcuts}~\cite{dancette2021beyond} and VQA-CP v2\footnote{https://www.iro.umontreal.ca/~agrawal/vqa-cp/}~\cite{agrawal2018don} datasets.

VQA-CE~\cite{dancette2021beyond} evaluation protocol is built on top of the VQA v2~\cite{goyal2017making} validation set. The Counterexamples subset consists of about 63K examples where all shortcuts lead to incorrect answers. The non-overlapping Easy subset has 147K examples where at least one shortcut points to the correct answer. The Hard subset includes 3K examples where no shortcuts exist or provide any clues for question answering. VQA-CP v2~\cite{agrawal2018don} is created by re-organizing the training and validation splits of the VQAv2 dataset. The distribution of answers per question type (`how many', `what color is', etc.) in test data is different from that in the training set. It has 438K and 220K questions respectively for training and testing.

We pre-train the METER~\cite{dou2021empirical} architecture on the COCO dataset as our Baseline model. Specifically, the vision encoder and the text encoder are CLIP~\cite{radford2021learning} and Roberta~\cite{liu2019roberta} respectively. The merged attention module consists of six layers of transformer blocks and a two-layer MLP block for the VQA head. To assess the efficacy concerning stronger baselines, we conduct experiments on the above datasets with the officially released METER\footnote{https://github.com/zdou0830/meter}~\cite{dou2021empirical} that is pre-trained on 4M data. Experiments on more architectures can be found in supplementary material. All the results we report are finetuned on the training split of each dataset for 10 epochs, with a batch size of 512 and a warm-up ratio of 0.1. We set $\lambda=0.5$ in the experiments with sensitivity analysis provided (in supplementary material), and $\tau=1.0$. To reduce training time, we use an image size of 224$\times$224 instead of the generally used 288$\times$288 or 576$\times$576 during finetuning, which takes about 1.5h/epoch on 4 NVIDIA V100 GPUs. 

\begin{table*}[]
    \centering
    \begin{tabular}{c|c|c|c|c|c|c}
        \hline
        \multirow{2}{4em}{Model} & \multicolumn{4}{c|}{VQA-CP v2 Scores} & \multicolumn{2}{c}{CLIP Scores} \\ \cline{2-7}
        & Yes/No & Other & Number & Overall & Positive$\uparrow$ & Negative$\downarrow$ \\ \hline
        Baseline & $42.47\pm 0.53$ & $\textbf{49.12}\pm 2.36$ & $15.68\pm 1.12$ & $42.02\pm 1.59$ & 0.264 & 0.082 \\
        Ours(Baseline) & $\textbf{44.72}\pm 0.22$ & $48.55\pm 1.07$ & $\textbf{18.21}\pm 0.87$ & $\textbf{42.80}\pm 0.40$ & \textbf{0.273} & \textbf{0.081} \\ \hline
        METER & $54.97\pm 0.21$ & $\textbf{64.26}\pm 0.03$ & $23.53\pm 0.24$ & $55.30\pm 0.07$ & 0.257 & \textbf{0.081} \\
        Ours(METER) & $\textbf{56.46}\pm 0.21$ & $63.73\pm 0.42$ & $\textbf{25.27}\pm 0.10$ & $\textbf{55.85}\pm 0.09$ & \textbf{0.273} & \textbf{0.081} \\
        \hline
    \end{tabular}
    \caption{Reasoning performance and CLIP scores of positive and negative sample pairs by baselines and ours on the VQA-CP v2 dataset. The best results are highlighted in \textbf{bold}.}
    \label{tab:baseline}
\end{table*}

\setlength{\tabcolsep}{0.1pt}
\begin{figure*}[ht]
    \centering
    \begin{tabular}{cc}
        \includegraphics[width=.46\textwidth]{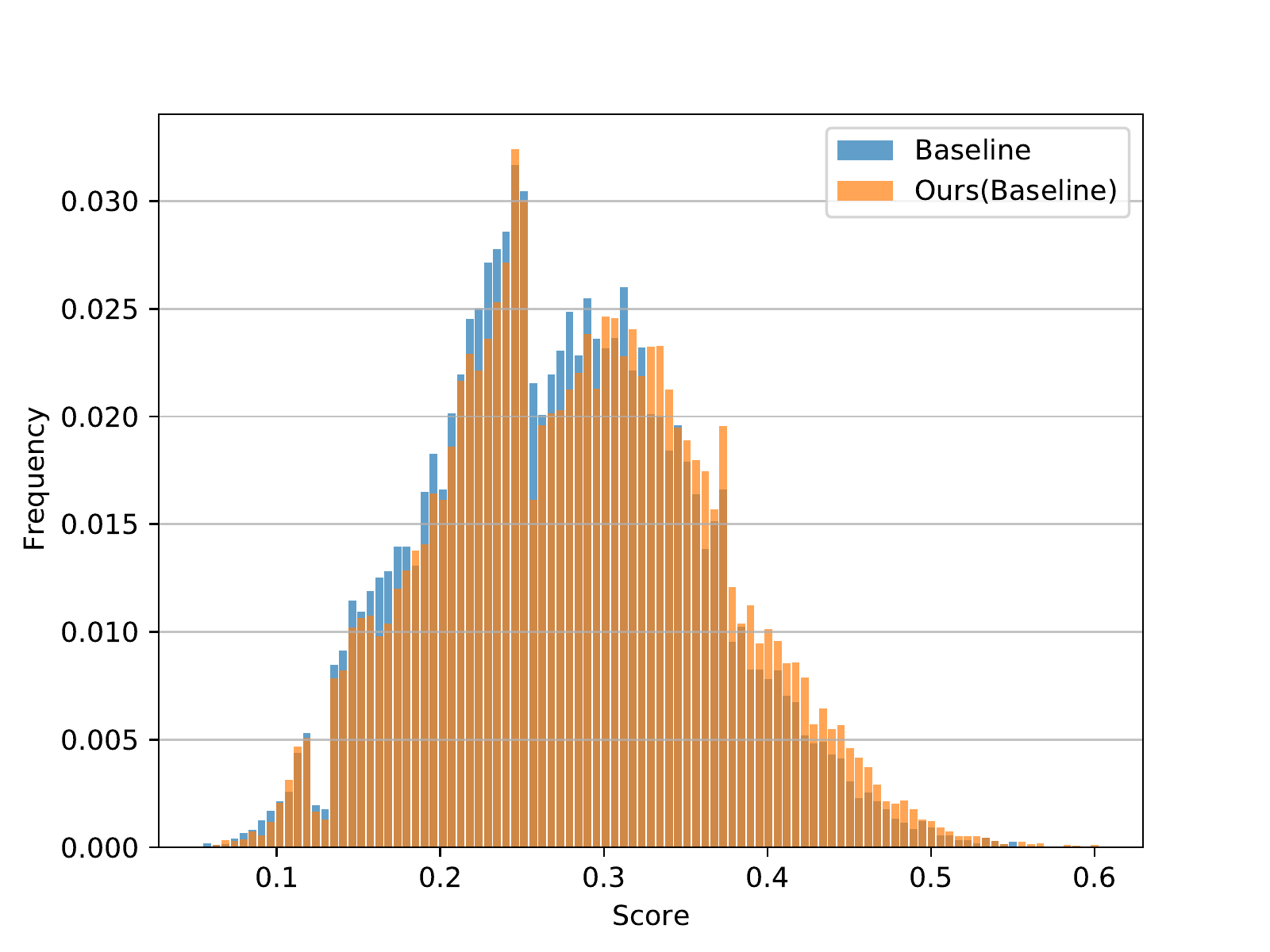} & \includegraphics[width=.46\textwidth]{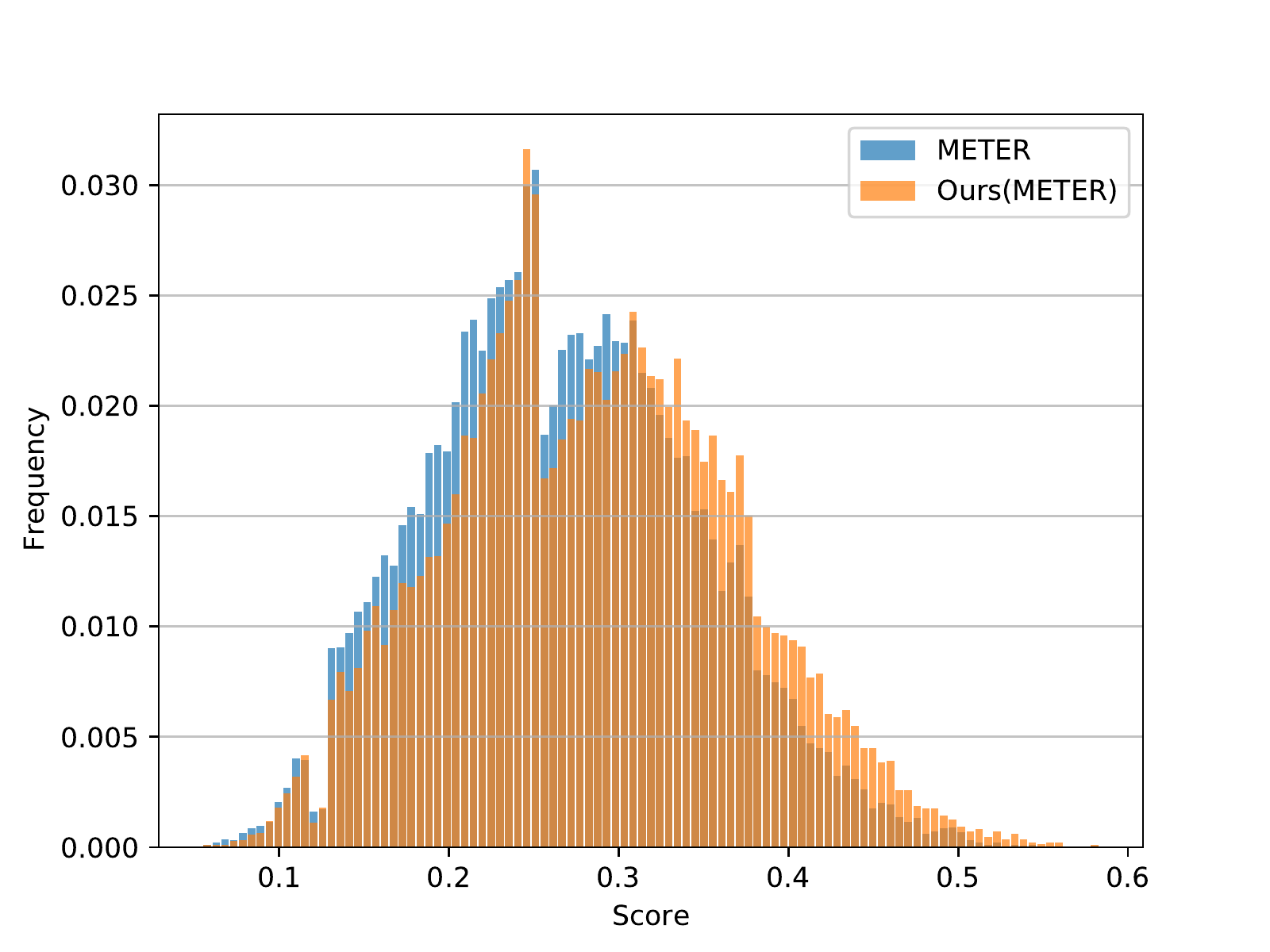} \\
        (a) & (b) 
    \end{tabular}
    \caption{The histograms of CLIP score for positive sample pairs by (a) the weak baseline and ours and (b) the strong baseline and ours. More results can be found in the supplementary material.}
    \label{fig:clip_score}
\end{figure*}

\begin{figure*}[th]
    \centering
    \includegraphics[width=.9\textwidth]{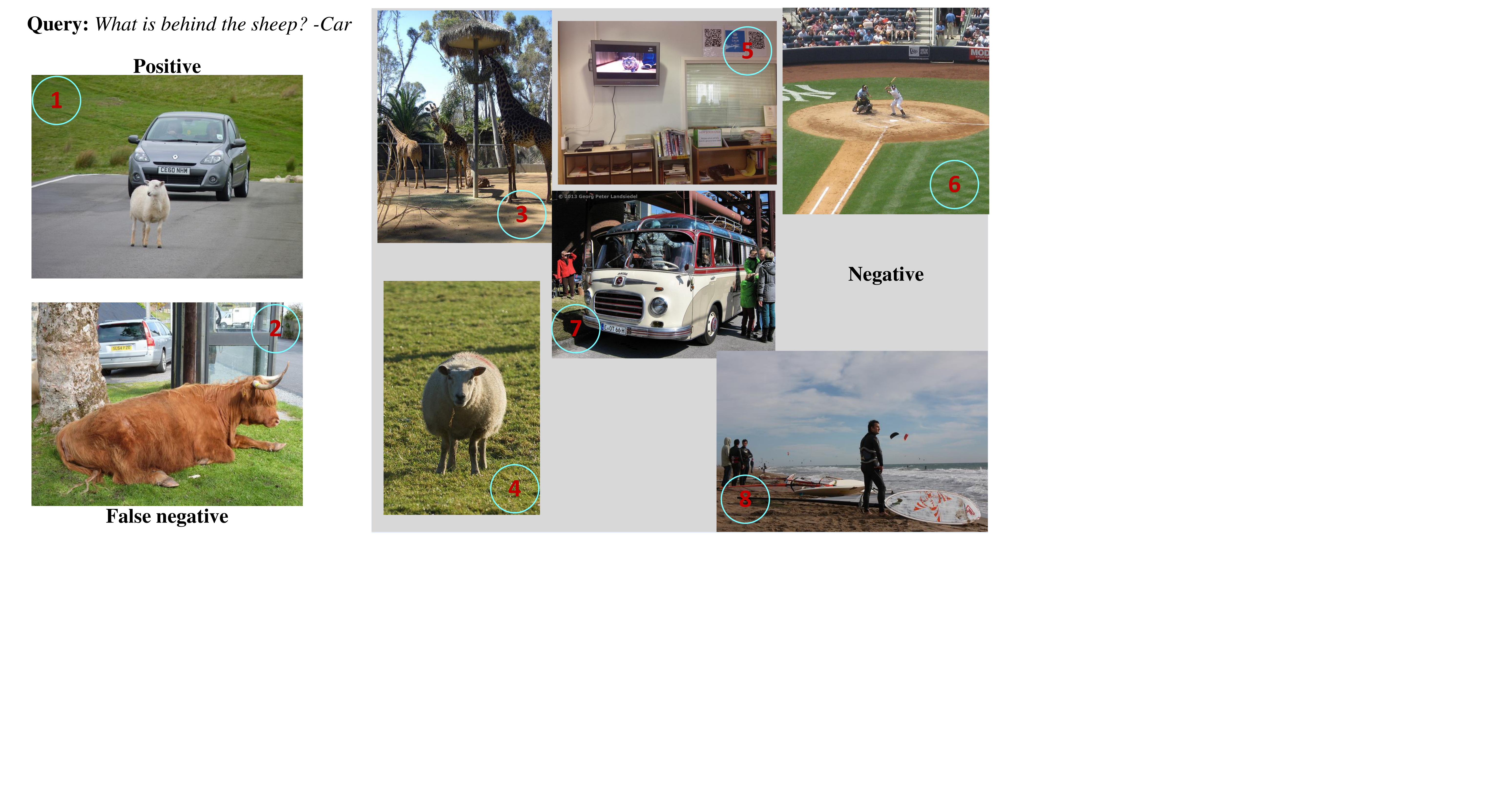}
    \caption{Positive and negative samples probing by our method (marked as \textbf{Positive}, \textbf{Negative} and \textbf{False negative} in the figure) and by supervised training from the same initialization. The supervised baseline identifies \textbf{\textcolor{red}{5}} and \textbf{\textcolor{red}{8}} as negative samples of the input query, and all the rest images as positive.}
    \label{fig:finer}
\end{figure*}

\subsection{Comparison with Baselines}

\paragraph{Quantitative comparison.}
We first compare the proposed method with the weak and strong baselines on the VQA-CP v2 dataset. We repeat the experiments for five times randomly and report question answering accuracy with $\bar{x}\pm{\sigma}$ in Table~\ref{tab:baseline}. Our method performs better on the challenging \textit{Yes/No} and \textit{Number} categories that are more sensitive to the shortcut issue, and slightly decreases on \textit{Other} questions. We hypothesize that \textit{Other} category benefits more from multi-modal shortcuts. The \textit{Overall} score is also improved. 

Since there is no ground truth of truly false negatives, we use the officially-released CLIP model\footnote{https://github.com/openai/CLIP} to compare the multi-modal alignments. The positive and negative scores in Table~\ref{tab:baseline} are calculated by $(I^+,QA^*)$, and $(I^-,QA^*)$, respectively. $(I^*,QA^*)$ are the annotated pairs, and $I^+,I^-$ are found by the VQA models. False negative samples are likely to lie in the overlap value range of positive and negative distributions (please see supplementary material for more details). Therefore, even with a low average negative score, it matters that the score distribution of positive samples moves toward a higher CLIP score in Figure~\ref{fig:clip_score}. Our method obtains higher positive scores and improved distributions, which indicates an enhanced ability to find true-positive samples.

\setlength{\tabcolsep}{7pt}
\begin{table*}[t]
    \centering
    \begin{tabular}{ccc|c|c|c|c|c|c}
    \hline
        \multicolumn{3}{c|}{Components} & \multicolumn{3}{c|}{VQAv2 test-dev} & \multicolumn{3}{c}{VQA-CE} \\ \hline
        CL & Finer-grained & Graph & Yes/No & Number & Other & Hard & Counter & Easy \\ \hline
        \xmark & \xmark & \xmark & 83.49 & 45.36 & \textbf{58.82} & 48.85 & \textbf{40.74} & 78.03 \\ 
        \cmark & \xmark & \xmark & 61.18 & 30.13 & 14.94 & 13.91 & 5.86 & 47.40 \\
        \cmark & \cmark & \xmark & 83.15 & 45.71 & 58.45 & 49.15 & 40.38 & 77.94  \\
        \cmark & \cmark & \cmark & \textbf{83.53} & \textbf{46.26} & 58.50 & \textbf{49.72} & \textbf{40.74} & \textbf{78.15} \\ \hline
    \end{tabular}
    \caption{Reasoning performance on VQAv2 test-dev split and VQA-CE dataset. The models are pre-trained on the COCO dataset, and finetuned on the VQAv2 training split. The key components include contrastive learning (CL), finer-grained sample organization (Finer-grained), and connected graphs (Graph). Highest scores are highlighted in \textbf{bold}.}
    \label{tab:ablation}
\end{table*}

\setlength{\tabcolsep}{10pt}
\begin{table*}[t]
    \centering
    \begin{tabular}{c|c|c|c|c|c|c}
    \hline
        \multirow{2}{4em}{Methods} &  \multicolumn{2}{c|}{VQAv2 test-std} & \multicolumn{3}{c|}{VQA-CE} & \multirow{2}{5em}{VQA-CP v2} \\ \cline{2-6} 
        & Overall & Number & Hard & Counter & Easy &  \\ \hline
        SANs~\cite{yang2016stacked} & 58.90 & 36.60 & 37.10 & 26.64 & 68.45 & 24.69 \\ 
        UpDown$^*$~\cite{anderson2018bottom} & 70.34 & 48.64 & 42.57 & 33.91 & 76.69 & 39.74 \\
        BLOCK~\cite{ben2019block} & 66.41 & 44.76 & 42.82 & 32.91 & 77.65 & 38.69 \\
        VilBERT~\cite{lu2019vilbert} & 70.92 & - & 45.82 & 39.24 & 80.50 & - \\
        ConClaT~\cite{kant2021contrast} & 68.77 & 49.99 & 49.39 & 41.05 & 79.70 & - \\ \hline
        Baseline & 67.64 & 45.39 & 48.85 & 40.74 & 78.03 & 42.02 \\
        Ours(Baseline) & 67.80 & \textbf{46.21} & \textbf{49.72} & 40.74 & 78.15 & \textbf{42.80} \\
        METER~\cite{dou2021empirical} & 74.11 & 51.84 & 60.01 & 51.69 & 84.41 & 55.30 \\
        Ours(METER) & 74.21 & \textbf{52.66} & 60.00 & 51.70 & 84.48 & \textbf{55.85} \\ \hline
    \end{tabular}
    \caption{Comparison on VQAv2 test-std split, VQA-CE, and VQA-CP v2 datasets. UpDown$^*$ is an ensemble of 30 models. For the VQAv2 test-std, the results of ConClaT, Baseline, METER, and Ours are finetuned on the training split. Improvements over 0.5\% compared with baseline models are highlighted in \textbf{bold}.}
    \label{tab:compare}
\end{table*}

\paragraph{Qualitative comparison.}
To have a more intuitive comprehension of the proposed cross-modal contrastive learning, we compare finer-grained contrastive samples learned by our method, i.e., Ours(METER), and that identified by supervised training, i.e., METER. After finetuning on the VQA-CP v2 dataset, we build connected graphs for a mini-batch of images. We list the first eight images in Figure~\ref{fig:finer} and show the identification of our method. Interestingly, we observe that METER gives the same discrimination result for these images before and after the finetuning. It seems that the model simply classifies them according to visual similarity, e.g., grassland, vehicles, animals, etc. It is worth noting that Ours(METER) and METER have the same initialization. Our method that jointly trains with finer-grained cross-modal contrastive learning does learn soft alignment between vision and language modalities, which verifies our hypothesis at the beginning. Please see the supplementary material for reasoning examples in different cases.

\subsection{Ablation Studies}
To verify the efficacy of the key components in our method, we conduct ablation studies on VQAv2 and VQA-CE datasets. The Baseline that has none of the components serves as a reference. The variant that exploits contrastive learning without finer-grained contrastive samples or connected graphs mimics the joint contrastive training using image negatives in ConClaT~\cite{kant2021contrast}, where $(IQA, I^-QA)$ are negative pairs. The variant that removes connected graphs implements the joint contrastive learning in Eq.~(\ref{eq:base}).

From the results in Table~\ref{tab:ablation}, we see that coarse-level contrastive learning interferes with the original supervised training and suffers from a heavy performance drop in all cases. ConClaT~\cite{kant2021contrast} addresses this issue by designing a large number of text-augmented negative samples. Alternatively, we target the problem with the proposed finer-grained sample organization, i.e., regarding matched question-answer pair as a whole. The variant equipped with contrastive learning and finer-grained samples shows improved reasoning ability on \textit{Number} and \textit{Hard} questions but worse performance on other categories such as \textit{Yes/No} and \textit{Counter}. Benefiting from the connected graphs that reduce false negative samples, our method succeeds in maintaining the performance on most shortcut-related categories and further promotes the accuracy on challenging questions, i.e., 0.55\% on \textit{Number} questions and 0.57\% on \textit{Hard} questions.

\subsection{Comparison with Previous Results}
We compare the proposed method with several representative VQA approaches. SANs~\cite{yang2016stacked} include a multi-layer stacked attention network that queries the grid features of an image multiple times to infer answers. UpDown~\cite{anderson2018bottom} estimates soft object-level attention at each word step and obtains the final prediction. BLOCK~\cite{ben2019block} applies bilinear fusion on question embedding and image embedding to conduct one-stage reasoning. VilBERT~\cite{lu2019vilbert} pre-trains a two-stream BERT extension to learn multi-modal representations, and finetune a two-layer MLP upon the element-wise product of objects and text representations for VQA. ConClaT~\cite{kant2021contrast} builds on top of Uniter~\cite{chen2020uniter} and introduces three types of contrastive samples to alternately train with supervised VQA annotations. METER~\cite{dou2021empirical} pre-trains uni-modal encoders and multi-modal fusion layers, and finetune an MLP block for reasoning with supervised learning. We adopt the basic architecture similar to METER as our Baseline, which pre-trains on the COCO dataset and is finetuned by supervised training. Among the approaches, SANs~\cite{yang2016stacked}, Baseline, METER~\cite{dou2021empirical} and our method learn grid features for images, UpDown~\cite{anderson2018bottom}, BLOCK~\cite{ben2019block}, VilBERT~\cite{lu2019vilbert} and ConClaT~\cite{kant2021contrast} exploit object detection for image representations.


Experimental results are listed in Table~\ref{tab:compare}. First, we evaluate models on the widely-used VQAv2 dataset to obtain an averaged performance. The results of comparison methods are from either their papers or released prediction files and are given by the official evaluation server. We can see that the overall reasoning accuracy of different methods is much higher than that for \textit{Number} questions. The object-based approaches including UpDown~\cite{anderson2018bottom}, BLOCK~\cite{ben2019block}, VilBERT~\cite{lu2019vilbert} and ConClaT~\cite{kant2021contrast} show an obvious advantage over SANs~\cite{yang2016stacked} and the Baseline. The strong baseline METER~\cite{dou2021empirical} promotes the performance by a large margin. For either the Baseline or METER, our method shows a consistent $\sim$0.8\% increase on \textit{Number} accuracy and a slight improvement in overall accuracy. It demonstrates that our method is more robust to challenging questions where shortcuts are less likely to be utilized, and is insensitive to the strength of baseline models.

Compared with the VQAv2 dataset which classifies questions according to their linguistic types, the VQA-CE dataset categorizes questions by whether involving shortcuts. Easy questions can be correctly answered using shortcuts, and counterexamples are likely to be answered reversely using shortcuts. So there is an implicit trade-off between the reasoning performance on easy and counterexamples. Compared with UpDown~\cite{anderson2018bottom}, BLOCK~\cite{ben2019block}, which builds upon the UpDown model, increases Easy accuracy by $\sim$1.0\% at the cost of a decrease on Counter questions by $\sim$1.0\%, and slightly promotes the performance on \textit{Hard} questions. Overall improvements in three categories are observed with the development of architectures. Compared with Baseline, our method achieves higher (0.87\%) accuracy on \textit{Hard} questions and maintains or slightly increases on the other two types. For the stronger METER model, our method keeps its impressive performance. A similar conclusion can be drawn from the VQA-CP v2 dataset, where shortcuts are explicitly removed for test questions. It is worth noting that even though the METER model has already achieved remarkable results on VQA-CP v2 test set, our method can still bring a considerable improvement (0.55\%).


\section{Conclusion}\label{sec:Con}
We proposed a finer-grained cross-modal contrastive learning method that jointly trains with a supervised learning objective for robust reasoning. Experiments in different settings are conducted to show that our method promotes reasoning ability in cases where shortcuts are less likely to handle. The proposed learning strategy encourages a soft alignment between vision and language modalities and extracts more abstract knowledge by distinguishing contrastive samples. In this paper, we present that joint training with a vanilla contrastive learning strategy contributes to robust reasoning. However, a main {\bf limitation} of this paper is that the latest techniques in contrastive learning are not involved and should be further explored. In future works, we hope that self-supervised methods can not only eliminate the shortcut issue but also improve the performance on all categories of questions.

\bibliographystyle{plain}
\bibliography{templateArxiv} 

{\appendix
\section{More Experimental Results}
\paragraph{CLIP Score distribution.} We present the full CLIP score\footnote{https://github.com/openai/CLIP} distribution in Figure~\ref{fig:clip_score_full} for a more intuitive comparison. Positive samples should score higher while negative samples lower. Since the CLIP evaluator rarely sees a question-answer text, the highest score of positive pairs is about 0.6, which is largely lower than 1.0. 

The score distributions of negative samples from different models approximate a Gaussian distribution with $\mu\approx 0.075$ and a range of (0, 0.175). The score distribution of positive samples has a range of (0, 0.6). False negative samples are likely to lie in the apparent overlap range of positive and negative distributions, e.g., (0.10, 0.15). Therefore, even with a low average negative score, it matters that the score distribution of positive samples moves toward a higher CLIP score in Figure~\ref{fig:clip_score_full}. Our method obtains higher positive scores and improved distributions, which indicates an enhanced ability to find true-positive samples.

\paragraph{Sensitivity analysis.} For the weighting hyperparameter in Eq.~(\ref{eq:joint}), we conduct sensitivity analysis on VQAv2 and VQA-CE datasets and present the results in Figure~\ref{tab:lambda}. We test the reasoning accuracy at $\{0,~0.1,~0.3,~0.5,~0.7,~1.0\}$. Consistent trends can be observed in the two datasets. The performance on questions that are more likely to be answered by shortcuts such as \textit{Yes/No} and \textit{Easy} types are insensitive to the change of $\lambda$ when it varies in $[0,~0.7]$. For the remaining more challenging questions, higher scores are achieved by relatively larger $\lambda$, e.g., $0.5$ or $0.7$. The results verify our hypothesis that joint contrastive learning mainly contributes to robust reasoning where the shortcut is less likely to help and influences little on superficial reasoning.

\setlength{\tabcolsep}{0.1pt}
\begin{figure*}[ht]
    \centering
    \begin{tabular}{cc}
        \includegraphics[width=.46\textwidth]{figs/baseline_pos_score.pdf} & \includegraphics[width=.46\textwidth]{figs/meter_pos_score.pdf} \\
        (a) & (b) \\
        \includegraphics[width=.46\textwidth]{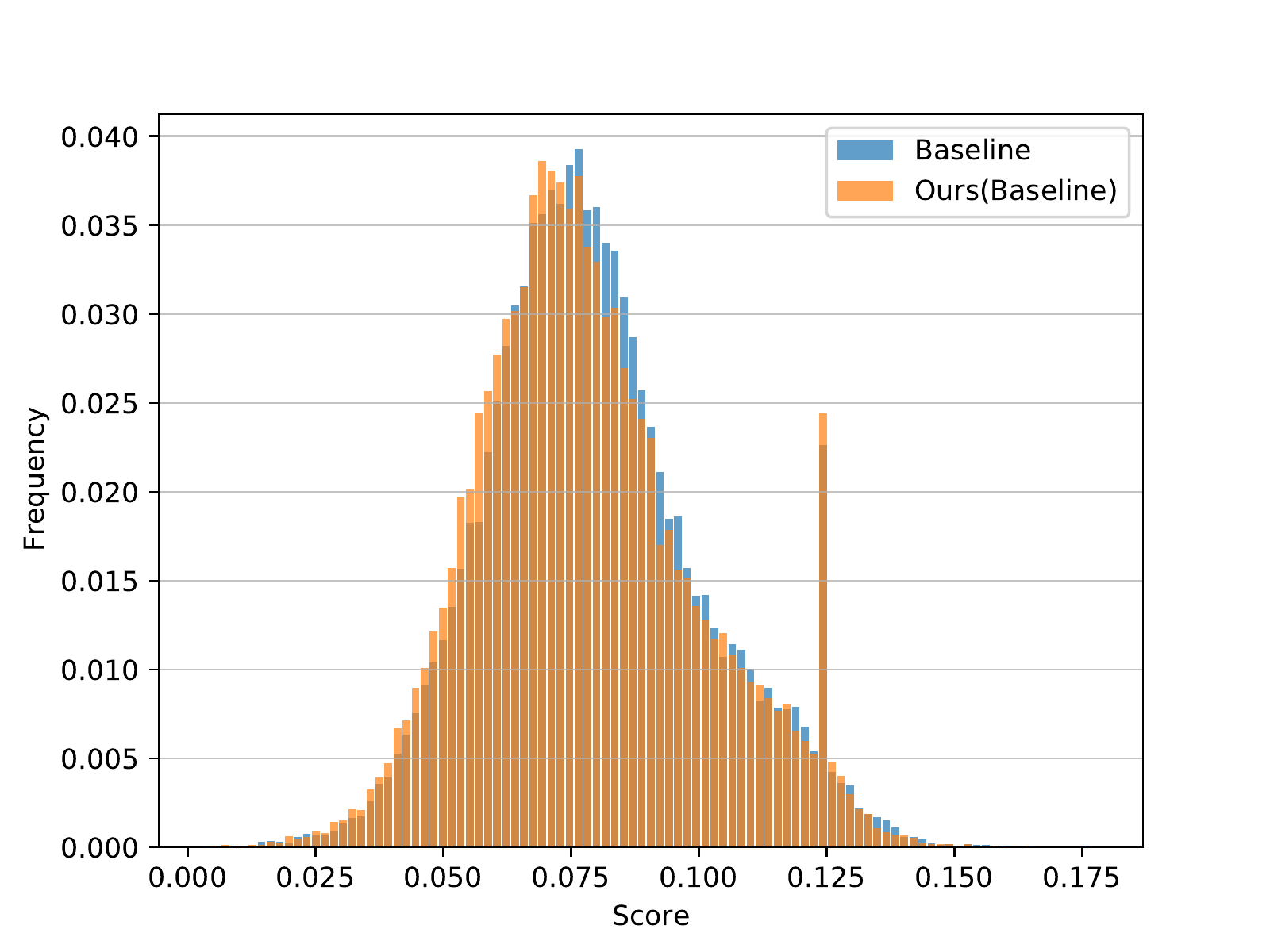} & \includegraphics[width=.46\textwidth]{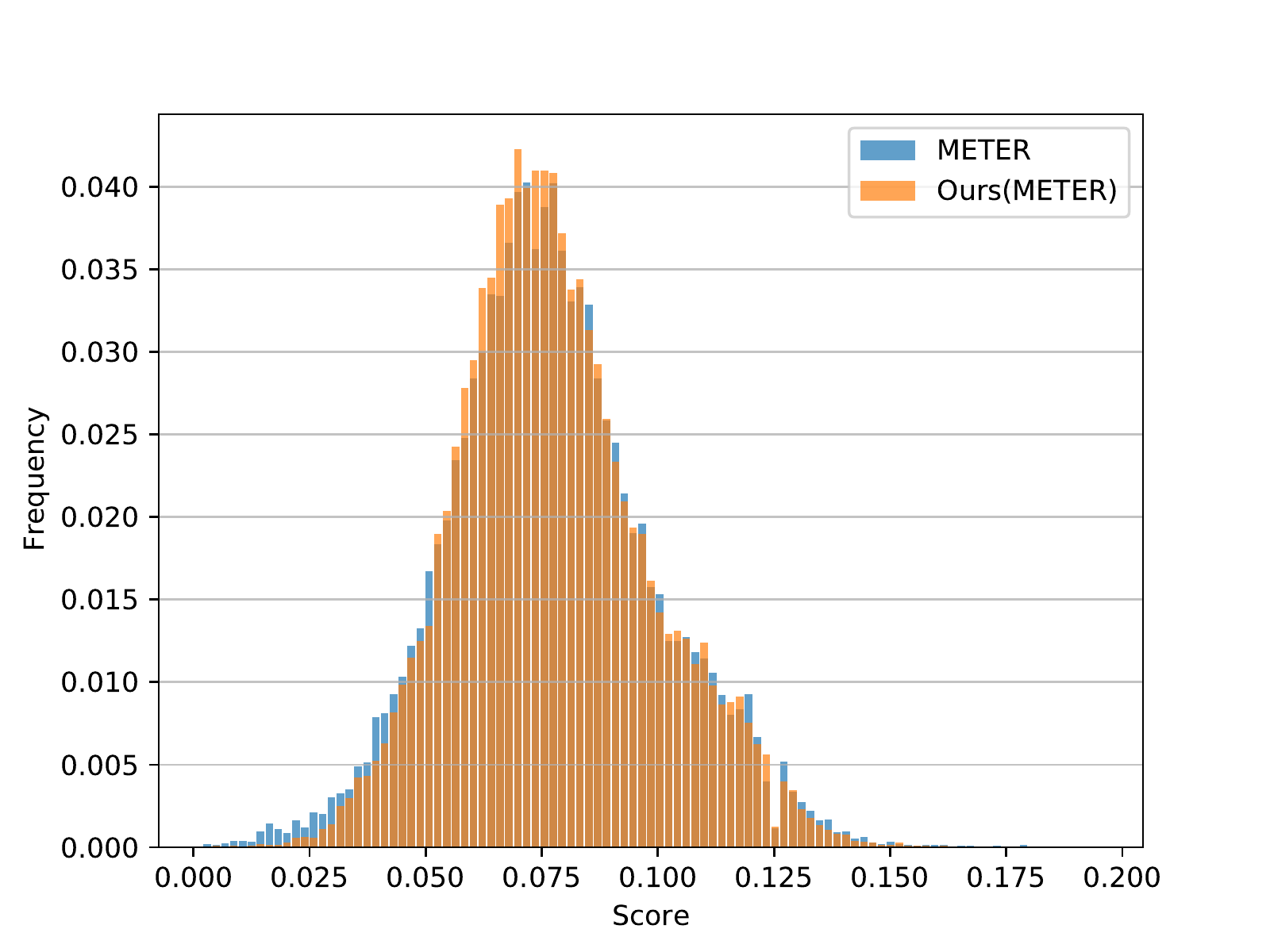} \\
        (c) & (d)
    \end{tabular}
    \caption{The histograms of CLIP score for positive sample pairs by (a) the weak baseline and ours and (b) the strong baseline and ours, and for negative sample pairs in (c) and (d).}
    \label{fig:clip_score_full}
\end{figure*}

\paragraph{Results on different architectures.}
To verify the efficacy of our method, we conduct experiments with the ALBEF~\cite{li2021align} backbone architecture. The result in Table~\ref{tab:albef} supports the conclusion that our method improves reasoning on shortcut-sensitive \textit{Yes/No} questions and challenging \textit{Number} questions. The promotion is slightly weaker than that in METER~\cite{dou2021empirical} architecture. We hypothesize that the image-text contrastive loss used during pre-training has already increased the robustness of ALBEF. However, our method further boosts the performance in case of heavily biased distribution, e.g., the \textit{Yes/No} category, which verifies the significance of contrastive learning during finetuning.

\setlength{\tabcolsep}{1.7pt}
\begin{table}[!h]
    \centering
    \begin{tabular}{c|c|c|c|c}
        \hline
        Model & Yes/No & Other & Number & Overall \\ \hline
        ALBEF & 48.36$\pm$0.62 & \textbf{52.00}$\pm$0.06 & 18.09$\pm$0.17 & 45.76$\pm$0.17 \\
        Ours & \textbf{51.16}$\pm$0.23 & 50.72$\pm$0.04 & \textbf{18.33}$\pm$0.08 & \textbf{45.93}$\pm$0.09 \\ \hline
    \end{tabular}
    \caption{Comparison with ALBEF architecture on VQA-CP v2 dataset. Highest scores are highlighted in \textbf{bold}.}
    \label{tab:albef}
\end{table}

\paragraph{VQA in different cases.} Figure~\ref{fig:vis_vqa} list some reasoning examples in test data. The first row mainly queries specific items. From the results, we see that the Baseline tends to give inexact answers, e.g., \textit{bread} to the first question, and is more likely to answer by correlation, e.g., \textit{phone} for \textit{person} and \textit{glasses} for \textit{desk}. Our method predicts relatively correct answers, and the performance increases with a stronger baseline model. For \textit{number} questions in the second row, Ours(Baseline) and Ours(METER) show improved robustness to the distractions in images, e.g., \textit{snow} for \textit{polar bear} and \textit{sky} for \textit{clouds}. Examples in the third-row query about precise locations. In these cases, our methods can target the related regions in an image and answers that are more consistent with human cognition.

\setlength{\tabcolsep}{0.1pt}
\begin{figure*}[ht]
    \centering
    \begin{tabular}{cc}
        \includegraphics[width=.42\textwidth]{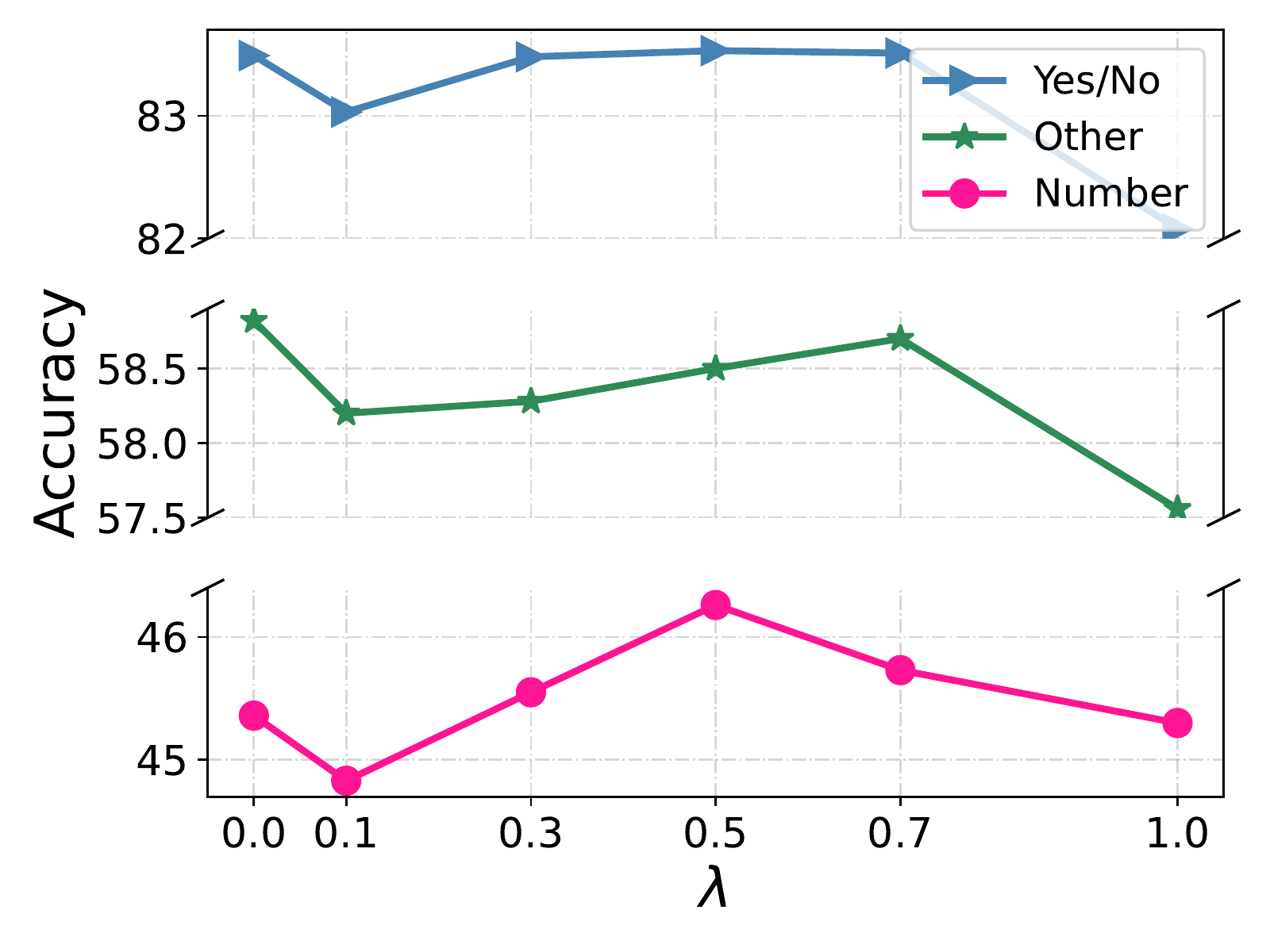} & \includegraphics[width=.42\textwidth]{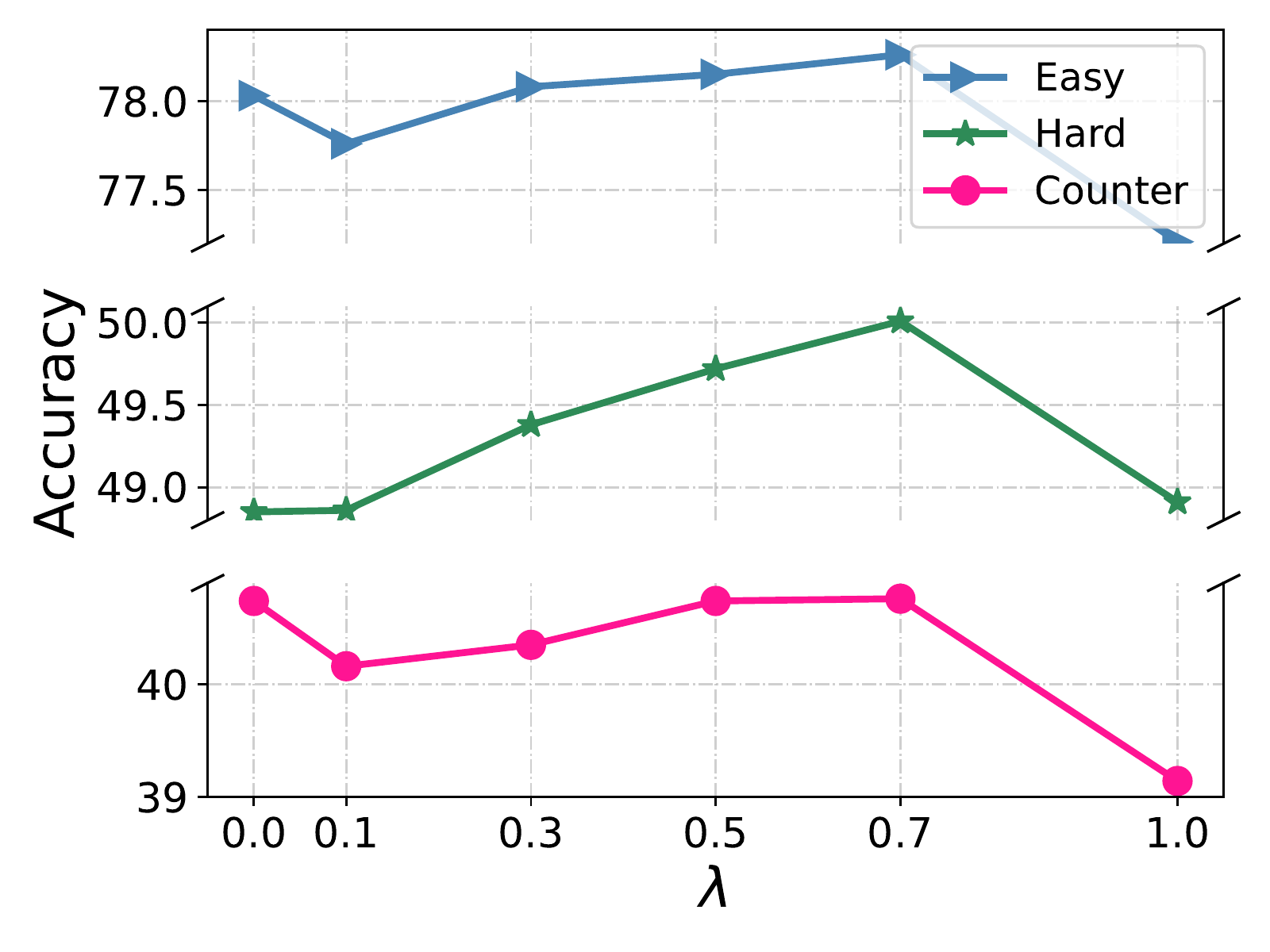} \\
        (a) & (b) 
    \end{tabular}
    \caption{Sensitivity of VQA accuracy w.r.t. $\lambda$ on (a) VQAv2 test-dev and (b) VQA-CE datasets.}
    \label{tab:lambda}
\end{figure*}

\begin{figure*}[t]
    \centering
    \includegraphics[width=.9\textwidth]{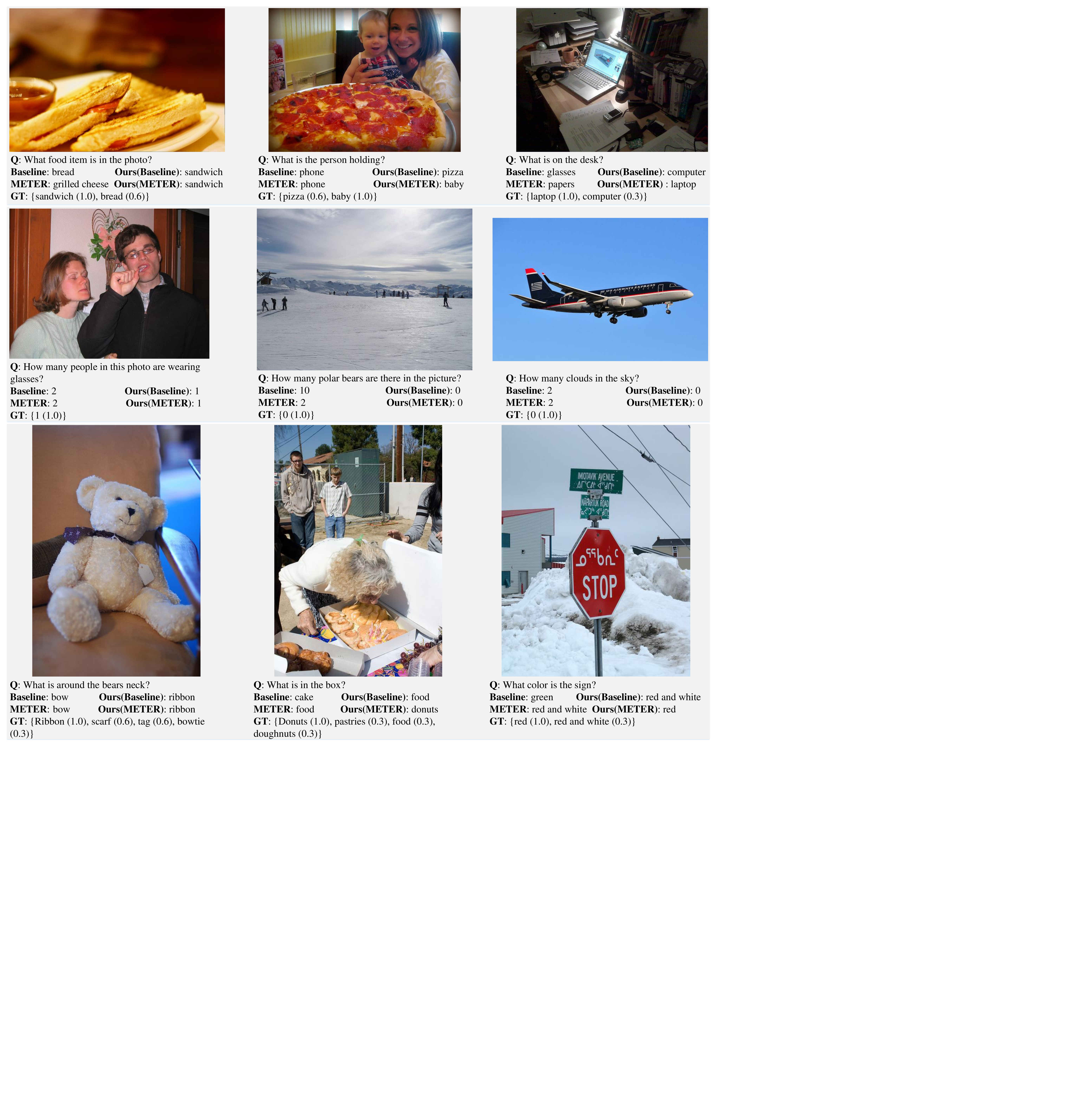}
    \vspace{-0.3cm}
    \caption{Reasoning examples on specific items, numbers and locations. Multiple answers in GT are distinguished by their frequency. For instance, \textit{sandwich (1.0), bread (0.6)} indicates that more than three human annotators answered \textit{sandwich} and two answered \textit{bread}.}
    \label{fig:vis_vqa}
\end{figure*}
}

\end{document}